\documentclass[10pt,twocolumn,letterpaper]{article}

\usepackage{cvpr}              
\usepackage{multirow}
\usepackage{booktabs}
\usepackage{makecell}
\usepackage{amssymb}
\usepackage{xcolor}
\usepackage{bbm}
\usepackage{algorithm}
\usepackage{algorithmic}
\usepackage[table]{xcolor}

\definecolor{cvprblue}{rgb}{0.21,0.49,0.74}
\usepackage[pagebackref,breaklinks,colorlinks,allcolors=cvprblue]{hyperref}

\def\paperID{8} 
\def\confName{CVPR}
\def\confYear{2026}

\title{Dual-Imbalance Continual Learning for Real-World Food Recognition}

\author{Xiaoyan Zhang\\
University of Michigan\\
Ann Arbor, Michigan, U.S.A\\
{\tt\small xyzaxis@umich.edu}
\and
Jiangpeng He\\
Indiana University\\
Bloomington, Indiana, U.S.A\\
{\tt\small jhe2@iu.edu}
}

\begin{document}
\maketitle
\begin{abstract}

Visual food recognition in real-world dietary logging scenarios naturally exhibits severe data imbalance, where a small number of food categories appear frequently while many others occur rarely, resulting in long-tailed class distributions. 
In practice, food recognition systems often operate in a continual learning setting, where new categories are introduced sequentially over time. 
However, existing studies typically assume that each incremental step introduces a similar number of new food classes, which rarely happens in real world where the number of newly observed categories can vary significantly across steps, leading to highly uneven learning dynamics. 
As a result, continual food recognition exhibits a \emph{dual imbalance}: imbalanced samples within each food class and imbalanced numbers of new food classes to learn at each incremental learning step.
In this work, we introduce \textbf{DIME}, a \textbf{D}ual-\textbf{I}mbalance-aware Adapter \textbf{Me}rging framework for continual food recognition. 
DIME learns lightweight adapters for each task using parameter-efficient fine-tuning and progressively integrates them through a class-count guided spectral merging strategy. 
A rank-wise threshold modulation mechanism further stabilizes the merging process by preserving dominant knowledge while allowing adaptive updates. 
The resulting model maintains a single merged adapter for inference, enabling efficient deployment without accumulating task-specific modules.
Experiments on realistic long-tailed food benchmarks under our step-imbalanced setup show that the proposed method consistently improves by more than $3\%$ over the strongest existing continual learning baselines. Code is available at \url{https://github.com/xiaoyanzhang1/DIME}.
\end{abstract}
    
\vspace{-0.6cm}
\section{Introduction}
\label{sec:intro}
\vspace{-0.2cm}
Food recognition plays an important role in real-world applications, including dietary monitoring, health assessment, and medical nutrition management \cite{liu2016deepfood,11275914}. In practical systems such as mobile dietary logging or clinical nutrition platforms, food images are continuously collected from diverse users and environments \cite{dalakleidi2022applying,rouhafzay2025image}. As new cuisines, brands, and preparation styles emerge over time, recognition models must continuously incorporate new knowledge while preserving previously learned categories. These requirements naturally motivate the study of continual learning for food recognition, where models learn from sequential data streams without catastrophic forgetting.

Recent works have explore continual learning in the food domain. Early studies~\cite{he2021online,tran2024class,yang2024learning} investigate incremental food classification under streaming data settings, enabling models to expand their recognition as new food categories appear. However, a key characteristic of real-world dietary data is that food consumption follows a highly long-tailed distribution. A small number of popular foods dominate daily intake, while many food categories appear only rarely. To better reflect this phenomenon, recent research introduced the long-tailed Food101, Visual Food Recognition (VFN) benchmark~\cite{he2023long} and its population-specific extensions, VFN186-Insulin and VFN186-T2D, which capture dietary patterns from different health conditions and user groups~\cite{11275914}. These datasets highlight the importance of addressing class imbalance in food recognition and motivate the study of long-tailed continual learning, where models must sequentially learn under skewed class distributions.

\begin{figure}[t]
    \centering
    \includegraphics[width=\linewidth]{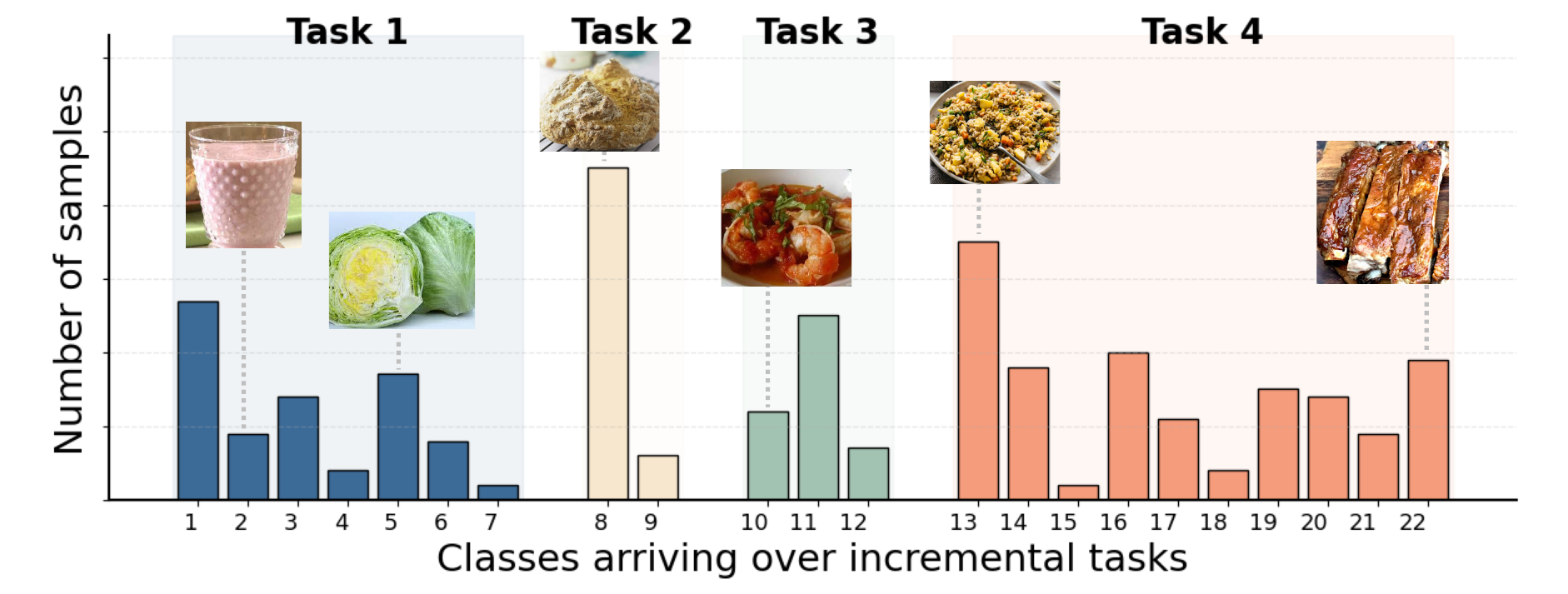}
    \vspace{-0.7cm}
    \caption{Illustration of the dual-imbalanced continual food recognition scenario. 
Tasks contain different numbers of classes (step imbalance), and each class has varying numbers of samples (long-tailed class distribution~\cite{liu2022longtailedclassincrementallearning}). 
    }
    \label{fig:intro}
    \vspace{-0.6cm}
\end{figure}

Despite these advances, an important aspect of real-world food data remains underexplored. Existing continual learning protocols typically assume balanced task structures, where each step contains a similar number of categories and follows a uniform learning schedule, although recent work has begun to explore more general step-imbalanced settings~\cite{zhang2026one}. Such settings simplify training dynamics but deviate from practical food streams, which are often highly irregular. In real-world scenarios, food datasets are inherently long-tailed at the class level, with a small number of frequent categories dominating the data while many others remain rare. Moreover, when constructing incremental tasks, categories are introduced over time in a non-uniform manner, producing substantial variation in class distributions across steps. As a result, real-world continual food recognition exhibits a \emph{dual imbalance}: (i) class imbalance, characterized by long-tailed class distributions across the whole dataset, and
(ii) step imbalance, where the number of classes varies across incremental steps, as illustrated in Figure~\ref{fig:intro}. While prior studies in food continual learning mainly address the former, the joint effect of these two imbalance factors remains largely unexplored. In this work, we study continual food recognition under this more realistic dual-imbalanced learning scenario. The main challenge arises from asymmetric learning dynamics caused by skewed data distributions: head food categories and large steps provide relatively stable gradients, whereas tail categories and small steps generate noisy and high-variance updates that easily disrupt previously learned representations.

To address this issue, we propose a lightweight and parameter-efficient continual learning framework tailored for food data streams. 
To mitigate the bias caused by long-tailed food distributions, we adopt Balanced Softmax training~\cite{ren2020balanced} as the base learning objective. Building upon this, we introduce a class-count aware spectral merging strategy that integrates newly learned knowledge with historical representations in a controlled manner. 
During merging, a rank-wise threshold modulation mechanism selectively preserves stable knowledge derived from head classes and large steps while allowing adaptive updates from small steps. The resulting model maintains a single merged adapter for inference, ensuring efficient deployment without accumulating task-specific modules. We conduct extensive experiments on four realistic long-tailed food benchmarks, including VFN186-LT, VFN186-Insulin, VFN186-T2D~\cite{11275914}, and Food101-LT~\cite{he2023long}, under step-imbalance protocols that simulate irregular real-world data streams. Our approach achieves state-of-the-art performance across multiple evaluation metrics, including final accuracy ($A_T$), average accuracy ($\bar{A}$), and the weighted average accuracy ($\mathrm{w}\bar{A}$) introduced to better evaluate performance under step-imbalanced learning scenarios. Comprehensive ablation studies further validate the effectiveness of each proposed component, while additional analyses demonstrate the efficiency of the framework and highlight its practical potential for real-world continual food recognition systems.
Our contributions are summarized as follows:
\begin{itemize}
\item We formulate a realistic continual food recognition scenario with \emph{dual imbalance}, capturing both long-tailed class distributions across the dataset and imbalanced class counts across incremental steps.
\item We propose DIME, a parameter-efficient continual learning framework that integrates class-count guided adapter merging with rank-wise threshold modulation, enabling stable knowledge integration under dual imbalance while maintaining a single merged adapter for inference.
\item We conduct extensive experiments on long-tailed food datasets under step-imbalance protocols. The results show consistent improvements over strong baselines and validate the effectiveness of the proposed components.
\end{itemize}





\section{Related Work}
\vspace{-0.1cm}
\label{sec:related_works}

\subsection{Visual Food Recognition}
\label{subsec:food_recog}
Visual food recognition has evolved significantly from early hand-crafted feature methods~\cite{bossard2014food,kawano2013real,anthimopoulos2014food} to deep representation learning~\cite{zhou2019application,11275914,ma2024mfp3d}. While early methods focused on low-level color and texture, deep neural networks like CNNs~\cite{krizhevsky2012imagenet} and Vision Transformers (ViTs)~\cite{dosovitskiy2020image} now dominate by automatically learning high-level semantic features~\cite{liu2025deep}. Recent advancements further incorporate pretrained models and Parameter-Efficient Fine-Tuning (PEFT)~\cite{wu2022few,yao2024caloraify} to capture fine-grained dish-ingredient relationships.
Despite these advances, visual food recognition remains inherently challenging due to high inter-class similarity, where distinct categories may appear visually similar, and substantial intra-class variability caused by diverse cooking methods, ingredients, and plating styles \cite{11433453,bossard2014food,mao2021visual}. 
Moreover, food categories are continuously evolving as new dishes and dietary habits emerge, making continual learning a natural framework for scalable food recognition systems.

\vspace{-0.1cm}
\subsection{Class Incremental Learning}
\label{subsec:cil}
Class Incremental Learning (CIL)~\cite{zhou2024class} aims to enable models to learn from a sequence of tasks while retaining previously acquired knowledge, mitigating the problem of catastrophic forgetting~\cite{mccloskey1989catastrophic}. Early CIL methods can be broadly categorized into three groups: regularization-based~\cite{li2017learning,hou2019learning,aljundi2018memory}, rehearsal-based~\cite{rebuffi2017icarl,castro2018end,iscen2020memory,shin2017continual}, and architecture-based approaches~\cite{hu2023dense,chen2023dynamic,yan2021dynamically}. 
With the emergence of powerful pretrained models (PTMs)~\cite{dosovitskiy2020image,radford2021learning}, recent continual learning research increasingly leverages pretrained representations for improved efficiency and scalability~\cite{zhou2024continual,he2025visplay}. Instead of training models from scratch, PTM-based CIL typically adapts pretrained features through lightweight mechanisms while keeping the backbone largely frozen. \emph{Prompt-based methods} introduce learnable prompts to guide the pretrained model for new tasks without modifying the backbone parameters~\cite{wang2022learning,wang2022dualprompt,smith2023coda}. \emph{Adapter-based methods} insert lightweight modules such as LoRA layers or MLP adapters into transformer blocks to incrementally acquire task-specific knowledge while preserving the pretrained representation~\cite{zhou2024expandable,he2025cl,wang2025self,liang2024inflora,sun2025mos,wang2025integrating}. These approaches provide strong parameter efficiency and have become a dominant paradigm for PTM-based continual learning.

\subsection{Long-Tailed and Imbalanced Learning}
\label{subsec:long_tailed}
Real-world visual recognition tasks often follow a long-tailed distribution, where a small number of classes contain abundant samples while many others have only a few~\cite{zhang2023deep,zhang2025systematic}. Existing work can be broadly categorized into two groups. Re-sampling methods construct a more balanced training batch by adjusting the numbers of samples in minority classes or majority classes~\cite{chawla2002smote,estabrooks2004multiple,liu2008exploratory}. Re-weighting approaches balance the contribution of different classes by modifying the loss or gradient during training, such as Balanced Softmax~\cite{ren2020balanced} and Influence-balance loss~\cite{park2021influence}. 
Handling data imbalance is particularly important in food recognition. Dietary habits are inherently dynamic and vary across seasons, health conditions, and cultural contexts, leading to naturally skewed data distributions \cite{chen2024metafood3d,fujihira2023factors,gligoric2021formation}. Recent studies have begun exploring long-tailed food recognition and constructing realistic datasets that follow such distributions, including long-tailed food benchmarks such as VFN-LT~\cite{11275914}. 
Some works have further attempted to combine long-tailed learning with continual learning to address class imbalance in sequential settings~\cite{11275914,bhattacharyya2025continual,he2021online}. 
However, these approaches typically assume a relatively fixed task structure where each task contains a similar number of categories. 
In practice, both the number of samples within a class and the number of categories introduced across tasks may vary significantly. 
This motivates a more realistic setting that simultaneously considers two types of imbalance: class imbalance within the dataset and step imbalance across tasks.




\section{Setup: Dual-Imbalance Class-Incremental Learning}
\label{sec:setup}

We consider a class-incremental learning (CIL) scenario where the model learns from a sequence of tasks while retaining knowledge of previously observed classes. Unlike standard CIL settings, we focus on a more realistic data stream exhibiting \emph{dual imbalance}, which includes both class-level imbalance within the whole dataset and step-level imbalance across tasks.

Let $\{\mathcal{D}_1, \mathcal{D}_2, \dots \mathcal{D}_T\}$ denote a sequence of training datasets arriving in $T$ incremental steps. Each dataset $\mathcal{D}_t = \{(x_i, y_i)\}_{i=1}^{N_t}$ contains $N_t$ training samples, where $x_i$ is an input image and $y_i$ is its corresponding food category label. The set of classes introduced at step $t$ is denoted as $\mathcal{C}_t$, and the classes are mutually exclusive across steps: $\mathcal{C}_i \cap \mathcal{C}_j = \emptyset, \quad i \neq j$.
At step $t$, the learner only has access to the current dataset $\mathcal{D}_t$ and must update the model to recognize all classes $\mathcal{C}_{1:t} = \bigcup_{k=1}^{t} \mathcal{C}_k$ observed so far.
During inference, the model predicts labels over the entire accumulated class set $\mathcal{C}_{1:t}$ without knowing the task identity.

\vspace{-4mm}
\paragraph{Class Imbalance.}
Food recognition datasets exhibit a long-tailed distribution, where a small number of classes contain abundant samples while many others have only a few~\cite{11275914}. 
Such distributions are commonly observed in real-world datasets and have been explicitly modeled in long-tailed food recognition benchmarks.
Formally, let $n_c$ denote the total number of samples belonging to class $c$ in the entire dataset. 
In a long-tailed distribution, $n_{c_{\text{head}}} \gg n_{c_{\text{tail}}}$, where $c_{\text{head}}$ and $c_{\text{tail}}$ represent frequent and rare food categories, respectively. 
Note that this imbalance is defined over the global class distribution rather than within individual tasks. As a result, the sequence of tasks may inherit this skewed distribution as new classes are incrementally introduced.

\vspace{-5mm}
\paragraph{Step Imbalance.}
In addition to class-level imbalance, the number of classes introduced at each incremental step may also vary substantially. 
Let $|\mathcal{C}_t|$ denote the number of classes observed at step $t$. 
We refer to this phenomenon as \emph{step imbalance}, where task sizes follow a skewed distribution across the learning sequence~\cite{zhang2026one}.
To control the degree of imbalance, we introduce a step imbalance factor $\rho \in (0,1]$. 
We define a sequence of step proportions as $s_t = \rho^{\frac{t-1}{T-1}}, \quad t=1,\dots,T$, where earlier steps contain larger proportions and therefore more classes (i.e., $S_1 \ge S_2 \ge \dots \ge S_T$ when $\rho$ is small).
To avoid introducing an artificial curriculum where tasks always appear from head to tail, we randomly permute the task order to produce the final class sequence $|\mathcal{C}_1|, |\mathcal{C}_2|, \dots, |\mathcal{C}_T|$.
This procedure preserves overall imbalance while allowing tasks of different scales to appear at arbitrary positions in the learning stream.

Together, these two factors define the dual-imbalance setting studied in this work. 
Class imbalance arises from the global long-tailed distribution of categories, where $n_{c_{\text{head}}} \gg n_{c_{\text{tail}}}$. 
Step imbalance further introduces heterogeneity across incremental steps by allocating different numbers of classes to each task, where $S_t \propto s_t$. 
Consequently, both the class frequency distribution and the task sizes across the learning sequence become highly skewed, forming a more realistic continual learning scenario.

\section{Method}
\label{sec:method}

\begin{figure*}[t]
    \centering
    \includegraphics[width=1.0\linewidth]{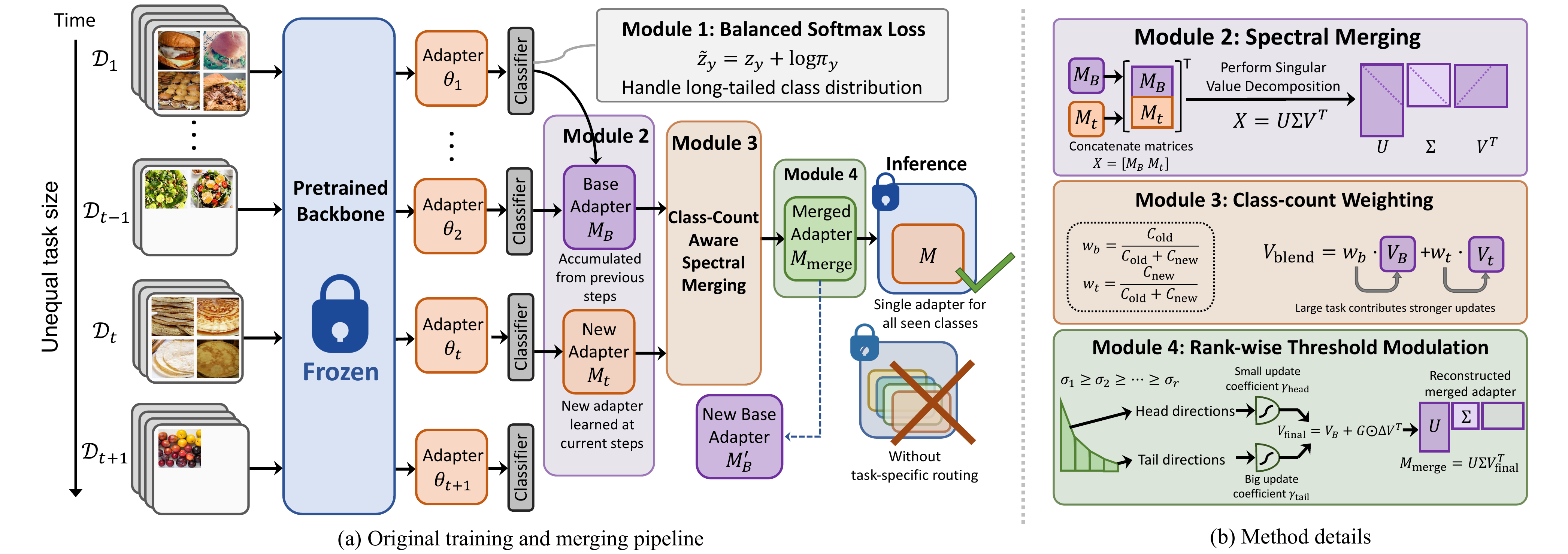}
    \vspace{-5mm}
    \caption{Overview of DIME. (a) Sequential food datasets are learned with lightweight adapters on a frozen backbone using Balanced Softmax. After each step, the new adapter is merged with the accumulated base adapter to produce a unified model. (b) The merging module performs spectral alignment via SVD, class-count guided weighting, and rank-wise threshold modulation.}
    \label{fig:framework}
    \vspace{-4mm}
\end{figure*}

Figure~\ref{fig:framework} presents the overall pipeline of the proposed framework.
Starting from a pretrained backbone, we attach a lightweight adapter that is trained for each learning step using Balanced Softmax described in Section~\ref{subsec:bsm} to handle long-tailed food supervision. 
After finishing a step, the newly trained adapter is integrated into the accumulated model through class-count aware spectral merging strategy as introduced in Section~\ref{sec:method_svd_merge}. 
A rank-wise threshold modulation mechanism in Section~\ref{sec:method_threshold} further modulates updates along singular directions to preserve stable knowledge while allowing flexibility for small tasks. 
The resulting merged adapter becomes the base model for subsequent steps and is used as the unified model during inference.

\subsection{Preliminary}
\label{subsec:pre}
To ensure parameter-efficient learning while mitigating the risk of catastrophic forgetting, we employ a pre-trained Vision Transformer (ViT) as the fixed feature extractor $\mathcal{B}$. In the context of visual food recognition, the backbone provides a rich, general-purpose representation of visual textures and structures. To specialize this knowledge for sequential food data, we incorporate lightweight bottleneck adapters into the Feed-Forward Network (FFN) of each Transformer block.

For a given input feature $x_{in}$ within a Transformer layer, the adapter module introduces a residual bottleneck transformation. The forward pass of the adapter can be mathematically formulated as follows:
\begin{equation}
    x_{out} = x_{in} + s \cdot \left( \mathbf{W}_{up} \cdot \sigma ( \mathbf{W}_{down} \cdot \text{LN}(x_{in}) ) \right)
\end{equation}
where $\text{LN}(\cdot)$ denotes Layer Normalization, and $\mathbf{W}_{down} \in \mathbb{R}^{d \times \hat{d}}$ and $\mathbf{W}_{up} \in \mathbb{R}^{\hat{d} \times d}$ are the down-projection and up-projection linear layers, respectively. To achieve significant parameter reduction, the bottleneck dimension $\hat{d}$ is set much smaller than the original embedding dimension $d$ (i.e., $\hat{d} \ll d$). The term $\sigma(\cdot)$ represents a non-linear activation function, and $s$ is a scaling factor that modulates the contribution of the adapter's residual branch.

In our sequential learning scenario, the pre-trained backbone parameters are kept frozen to preserve the fundamental visual knowledge required for food analysis. 
For each new task $t$ in the data stream, a task-specific set of adapter parameters $\theta_t$ is initialized. 
During the training phase of task $t$, only $\theta_t$ and the task-specific classifier head are optimized, while the backbone remains fixed.

This architectural design provides two important properties. 
First, the task-specific adapters enable rapid adaptation to newly introduced food categories and diverse cooking styles. 
Second, freezing the backbone preserves the general visual representations learned during pretraining, preventing representational drift across sequential tasks.

\subsection{Intra-task Balanced Optimization}
\label{subsec:bsm}
When training on sample-imbalanced streams, a standard Cross-Entropy (CE) loss tends to be dominated by these head categories~\cite{muhammad2025enhancing}. The model effectively learns a majority-rule bias, where it achieves low global loss by over-fitting to frequent items while essentially ignoring the rare food categories. This leads to a degradation in recognition performance for niche dietary items, which are often just as nutritionally relevant as staples.

To mitigate this bias and ensure a fair learning process across all food types, we adopt the Balanced Softmax strategy~\cite{ren2020balanced}. This approach explicitly incorporates class-frequency priors directly into the softmax computation to level the playing field. For a specific class $y$, given its predicted logit $z_y$ from the classifier, the adjusted logit $\tilde{z}_y$ is computed as:
\begin{equation}
    \tilde{z}_y = z_y + \log \pi_y
\end{equation}
In this formulation, $\pi_y$ denotes the empirical frequency of class $y$ within the current training task. By adding the log-prior $\log \pi_y$ to the raw logits, the loss function effectively penalizes over-confidence in head classes and encourages the model to assign higher importance to rare food categories during the optimization process.


\subsection{Class-Count Aware Spectral Merging}
\label{sec:method_svd_merge}

In the dual-imbalanced continual food recognition setting, different learning steps may contain drastically different amounts of information. Large steps often provide stable updates dominated by more food categories, while small steps introduce noisy gradients due to limited food classes. Directly merging model parameters from different steps can therefore lead to destructive interference. 

To address this issue, we adopt an alignment-then-merge strategy inspired by KnOTS~\cite{stoica2024model}. The key idea is to first align the representations of two adapters in a shared low-rank spectral space using singular value decomposition (SVD) and then perform merging in this space. While KnOTS was designed for LoRA updates, we extend the same principle to MLP-style adapter layers by applying the alignment procedure to each adapter parameter matrix.

\vspace{-4mm}
\paragraph{Spectral Merging.}
Let $M_B$ denote the parameter matrix of the base adapter accumulated from previous steps, and $M_t$ denote the adapter learned at the current step. 
Following the alignment idea of KnOTS~\cite{stoica2024model}, we first concatenate the two matrices along the column dimension
\begin{equation}
    X = [M_B \;\; M_t], \qquad
    X = U \Sigma V^\top .
\end{equation}
The decomposition defines a shared orthogonal basis $U$ that captures the principal directions of both adapters. 
In this basis, the coefficient matrix $V^\top$ contains the representations of $M_B$ and $M_t$ in the aligned coordinate system. 
Specifically, $V^\top$ can be split into two parts corresponding to the two adapters
\begin{equation}
    V^\top = [V_B^\top \;\; V_t^\top],
\end{equation}
where $V_B^\top$ and $V_t^\top$ represent the coordinates of $M_B$ and $M_t$ in the shared SVD space.
Compared with directly averaging parameters, performing merging in the aligned SVD space allows updates from different learning steps to interact along consistent principal directions, improving stability under highly imbalanced task sizes.

\vspace{-4mm}
\paragraph{Class-Count Weighting.}
After alignment, we combine the two representations in the shared spectral space. 
Let $C_{\text{old}}$ denote the number of previously learned classes and $C_{\text{new}}$ the number of newly introduced classes at the current step. 
We define merging weights
\begin{equation}
    w_b = \frac{C_{\text{old}}}{C_{\text{old}} + C_{\text{new}}}, \qquad
    w_t = \frac{C_{\text{new}}}{C_{\text{old}} + C_{\text{new}}}.
\end{equation}
Using these weights, we compute a blended representation
\begin{equation}
    V_{\text{blend}}^\top = w_b V_B^\top + w_t V_t^\top .
\end{equation}


\subsection{Rank-Wise Threshold Modulation}
\label{sec:method_threshold}


While the SVD alignment provides a shared representation space and the class-count guided rule determines the relative contribution of different steps, directly replacing $V_B^\top$ with $V_{\text{blend}}^\top$ may still lead to suboptimal knowledge integration. In particular, directions corresponding to dominant visual patterns should remain stable, while less significant directions should retain flexibility to accommodate newly emerging food categories.

To address this issue, we introduce the \emph{threshold modulation} that modulates the update applied along each singular direction according to its importance. 
After SVD decomposition, the singular values provide an ordered measure of energy in the aligned representation space. Directions associated with larger singular values typically capture stronger variations in food appearance, often reflecting prominent visual characteristics such as common color distributions, texture patterns, or coarse structural cues of dishes. Preserving these high-energy directions helps maintain stable representations for frequently observed foods. In contrast, directions with smaller singular values correspond to lower-energy components that can remain more adaptable to capture finer appearance variations, enabling the model to incorporate new visual patterns without disrupting previously learned representations.

We compute the update relative to the base adapter as $\Delta V^\top = V_{\text{blend}}^\top - V_B^\top$. 
Let $\Sigma = \text{diag}(\sigma_1,\dots,\sigma_r)$ denote the singular values obtained from the decomposition of the concatenated matrix. 
To regulate updates along different directions, we introduce a rank-based gating mask $G = [g_1,\dots,g_r]^\top$ where

\begin{equation}
g_i =
\begin{cases}
\gamma_{\text{head}}, & i \le r_h \\
\gamma_{\text{tail}}, & i > r_h ,
\end{cases}
\end{equation}
with $r_h$ denoting the head-rank threshold and $\gamma_{\text{head}}, \gamma_{\text{tail}}$ controlling the update strength for dominant and minor directions, respectively.
The final representation in the aligned SVD space is then computed as
\begin{equation}
V_{\text{final}}^\top = V_B^\top + G \odot \Delta V^\top ,
\end{equation}
where $\odot$ denotes element-wise multiplication applied row-wise.
Finally, the merged adapter is reconstructed in the original parameter space using the shared SVD basis

\begin{equation}
M_{\text{merge}} = U \Sigma V_{\text{final}}^\top .
\end{equation}

This mechanism is beneficial in the dual-imbalanced food learning scenario. 
Large steps usually produce stronger dominant directions in the SVD spectrum, while smaller steps contribute weaker but potentially useful variations. 
The proposed gating strategy stabilizes dominant structures while preserving sufficient flexibility for integrating information from rare food categories and small tasks.

\subsection{Training and Inference}
\label{sec:method_training}

At each learning step $t$, a task-specific adapter $\theta_t$ is trained on the current food dataset using the Balanced Softmax objective while keeping the pretrained backbone fixed. 
After training is completed, the newly learned adapter is integrated into the existing model using the proposed SVD alignment and class-count guided merging strategy with  threshold modulation. 
The merged result is stored as the updated base adapter, which accumulates knowledge across learning steps without increasing the number of parameters used during inference.
During inference, the model uses only the single merged adapter obtained from the sequential merging process. 
This design avoids storing or selecting multiple task-specific adapters and therefore provides a more efficient deployment compared with approaches that maintain separate adapters for each task.



\section{Experiments}
\label{sec:exp}

\subsection{Experimental Setups}
\label{subsec:exp_setup}

\paragraph{Datasets}
We evaluate the proposed method on four food recognition benchmarks: VFN186-LT, VFN186-Insulin, VFN186-T2D~\cite{11275914}, and Food101-LT~\cite{he2023long}. 
The VFN series datasets reflects real-world dietary consumption patterns across different populations and VFN186-Insulin and VFN186-T2D exhibit natural long-tailed class distributions. 
The distributions of food categories differ across these datasets, making them suitable for evaluating continual food recognition under realistic conditions. 
In addition, we include Food101-LT, a widely used long-tailed food benchmark derived from Food101~\cite{bossard2014food}, to further evaluate the generalization of our method.
Following the protocol in Section~\ref{sec:setup}, each dataset is partitioned into incremental steps with varying imbalance ratio $\rho$ to simulate step imbalance.

\vspace{-5mm}
\paragraph{Baselines.}
We compare the proposed method with several representative continual learning approaches, including SLCA~\cite{zhang2023slca}, FeCAM~\cite{goswami2023fecam}, InfLoRA~\cite{liang2024inflora}, EASE~\cite{zhou2024expandable}, SEMA~\cite{wang2025self}, CL-LoRA~\cite{he2025cl}, ACMap~\cite{fukuda2025adapter}, MOS~\cite{sun2025mos}, and TUNA~\cite{wang2025integrating}. 
These methods cover a variety of strategies such as parameter-efficient adaptation, prototype-based classification, and adapter merging for continual learning. 
All baselines are evaluated under the same experimental protocol for fair comparison.

\vspace{-5mm}
\paragraph{Implementation Details.}
Our implementation is based on the PyTorch framework. We adopt ViT-B/16~\cite{dosovitskiy2020image} pre-trained on ImageNet-21K~\cite{deng2009imagenet} as the backbone network. The model is trained for 20 epochs using the SGD optimizer with a learning rate of 0.07, weight decay of 0.0005, and a batch size of 16. For the adapter module, we employ an MLP adapter with a hidden dimension of 64. All experiments are conducted on a single NVIDIA L40S GPU.

\vspace{-5mm}
\paragraph{Evaluation Metrics.}
Following standard CIL evaluation protocols, we report the \emph{final accuracy} $A_T$, which measures the classification accuracy on all classes after the final learning step, and the \emph{average accuracy} $\bar{A}$ across all steps:
\[
\bar{A} = \frac{1}{T} \sum_{t=1}^{T} A_t ,
\]
where $A_t$ denotes the accuracy evaluated on the accumulated class set after completing step $t$.

Under step-imbalanced settings, however, the number of classes per task can vary substantially. In this case, earlier tasks with only a few classes
are typically easier and their $A_t$ can dominate the simple average $\bar{A}$.
As a complementary analysis, we additionally introduce the weighted average accuracy $\mathrm{w}\bar{A}$ to more fairly evaluate the overall performance under step-imbalance settings, where the number of classes per task varies within the same dataset and fixed task number $T$. 
It is defined as:
\begin{equation}
\mathrm{w}\bar{A}
= \frac{\sum_{t=1}^{T} w_t A_t}{\sum_{t=1}^{T} w_t},
\quad 
w_t = \frac{\sum_{i=1}^{t} |C_i|}{(Ct)/T},
\label{eq:weighted_acc}
\end{equation}
\noindent where $C = \sum_{i=1}^T |C_i|$ is the total number of classes. 
When $|C_i| = |C_j|$ for all $i,j \le t$, each $w_{t}=1$ and $\mathrm{w}\bar{A}$ reduces to the standard average accuracy $\bar{A}$~\cite{rebuffi2017icarl}. Intuitively, $\mathrm{w}\bar{A}$ assigns larger weights to stages where more classes have been accumulated, and thus provides an alternative view of overall performance under step-size imbalance.

\subsection{Main Results}
\label{subsec:main}

\begin{table*}[t]
\resizebox{0.95\textwidth}{!}{%
\begin{tabular}{l|ccc|ccc|ccc|ccc}
\Xhline{0.8pt}
Datasets
& \multicolumn{3}{c|}{\textbf{VFN186-LT}} 
& \multicolumn{3}{c|}{\textbf{VFN186-Insulin}} 
& \multicolumn{3}{c|}{\textbf{VFN186-T2D}} 
& \multicolumn{3}{c}{\textbf{Food101-LT}} \\
Evaluation Metrics
& $A_{T}$ & $\bar{A}$ & $\mathrm{w}\bar{A}$
& $A_{T}$ & $\bar{A}$ & $\mathrm{w}\bar{A}$
& $A_{T}$ & $\bar{A}$ & $\mathrm{w}\bar{A}$
& $A_{T}$ & $\bar{A}$ & $\mathrm{w}\bar{A}$ \\
\hline
SLCA~\cite{zhang2023slca} 
& 63.07 & 70.94 & 70.30 
& 63.10 & 73.11 & 71.59 
& 63.74 & 73.19 & 72.15 
& 72.35 & 80.25 & 79.53 \\


InfLoRA~\cite{liang2024inflora}       
& 51.49 & 62.10 & 60.99 
& 51.83 & 62.64 & 62.02
& 52.58 & 62.95 & 61.91
& 57.80 & 72.23 & 70.84 \\

EASE~\cite{zhou2024expandable}        
& 64.86 & 73.87 & 72.75
& 65.11 & 74.08 & 72.92
& 65.62 & 75.50 & 74.25 
& 68.00 & 78.10 & 77.01 \\

SEMA~\cite{wang2025self}              
& 46.22 & 58.78 & 57.68 
& 43.45 & 58.50 & 57.97 
& 45.17 & 60.16 & 59.69 
& 43.88 & 60.88 & 60.11 \\

CL-LoRA~\cite{he2025cl}               
& 60.40 & 72.53 & 71.68 
& 60.57 & 74.09 & 72.89 
& 58.84 & 72.83 & 71.19 
& 55.82 & 72.48 & 71.47 \\

ACMap~\cite{fukuda2025adapter}        
& 65.79 & 74.23 & \underline{73.46}
& 66.22 & 74.52 & \underline{74.58} 
& 67.07 & 75.18 & 74.23 
& 72.37 & 81.88 & 80.98 \\

MOS~\cite{sun2025mos}      
& 65.78 & 72.91 & 72.34
& 65.64 & 73.46 & 72.64
& 66.63 & 74.11 & 73.30 
& 74.31 & 81.74 & 80.78 \\

TUNA~\cite{wang2025integrating}      
& \underline{66.19} & \underline{74.42} & 73.43
& \underline{66.28} & \underline{74.97} & 73.88 
& \underline{67.32} & \underline{76.17} & \underline{74.29} 
& \underline{75.00} & \underline{82.60} & \underline{81.98} \\

\rowcolor{red!5}
DIME (Ours)
& \textbf{69.07} & \textbf{76.58} & \textbf{75.55} 
& \textbf{69.40} & \textbf{77.00} & \textbf{75.93} 
& \textbf{69.88} & \textbf{77.48} & \textbf{76.47} 
& \textbf{77.01} & \textbf{84.10} & \textbf{83.77} \\

\Xhline{0.8pt}
\end{tabular}
}
\centering
\vspace{-0.2cm}
\caption{Last task accuracy ($A_T$), average accuracy ($\bar{A}$), and weighted average accuracy ($\mathrm{w}\bar{A}$) on VFN186-LT, VFN186-Insulin, VFN186-T2D, and Food101-LT with $\rho=0.01$. Results are averaged over 5 runs. The best results are in \textbf{bold} and the second-best results are \underline{underlined}.}
\label{tab:main}
\vspace{-0.3cm}
\end{table*}

\begin{figure*}[t]
    \centering
    \includegraphics[width=1.0\linewidth]{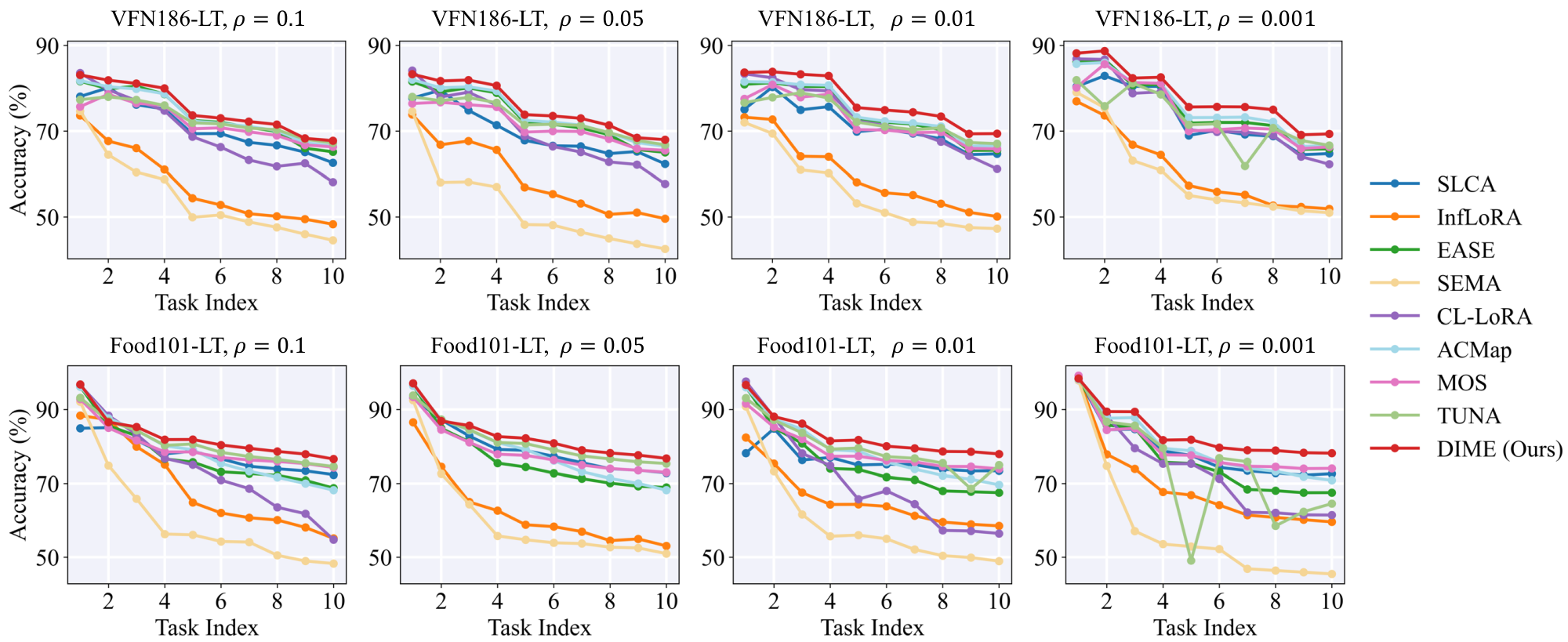}
    \vspace{-5mm}
    \caption{Performance comparison under different step-imbalance ratios $\rho$ on VFN186-LT and Food101-LT with $T=10$ incremental steps. Our method consistently achieves higher and more stable performance across tasks under all imbalance settings.}
    \label{fig:vfn_imbalance}
    \vspace{-5mm}
\end{figure*}

Table~\ref{tab:main} presents the overall performance comparison with competitive continual learning methods on four food recognition benchmarks with imbalance ratio $\rho=0.01$. DIME consistently achieves the best results across all evaluation metrics, including last-task accuracy ($A_T$), average accuracy ($\bar{A}$), and the weighted average accuracy ($\mathrm{w}\bar{A}$). 
On VFN186-LT, our method achieves an $A_T$ of $69.07\%$, outperforming the strongest baseline TUNA by $2.88\%$ and ACMap by $3.28\%$. On $\bar{A}$ and $\mathrm{w}\bar{A}$, DIME further achieves $76.58\%$ and $75.55\%$ respectively.
On the population-specific datasets VFN186-Insulin and VFN186-T2D, our approach also demonstrates consistent advantages. For example, on VFN186-Insulin, DIME improves $A_T$ with a gain of $3.12\%$ over the previous best method.
On the large-scale Food101-LT benchmark, DIME also demonstrates strong performance, achieving consistent improvements of around $2\%$ over the strongest baseline across all evaluation metrics.

Figure~\ref{fig:vfn_imbalance} further compares performance under different step-imbalance ratios $\rho$ on the VFN186-LT and Food101-LT. Across all imbalance settings, DIME consistently achieves the best accuracy throughout the learning sequence. Notably, the advantage becomes more pronounced when the imbalance becomes more severe. For example, under $\rho=0.001$, DIME achieves $69.33\%$ and $78.13\%$ after the final task on VFN186-LT and Food101-LT respectively, outperforming the strongest baseline by $2.73\%$ and $4.11\%$. 
In addition, the performance curves of DIME remain smoother and more stable across tasks. This trend becomes particularly evident when $\rho=0.001$, where the imbalance is most severe. Under this setting, some baseline exhibit pronounced fluctuations due to drastic variations in task sizes, whereas DIME maintains a much more stable trajectory. Consequently, DIME also achieves a higher $\mathrm{w}\bar{A}$, suggesting that the proposed merging and gating strategy effectively mitigates interference caused by highly imbalanced data streams.

\vspace{-5mm}
\paragraph{Task-Size Analysis}
To further analyze the effect of step-level imbalance, we report the performance on large, middle, and small tasks at the final step. Tasks are grouped according to the number of classes contained in each step, and we report the class-weighted average accuracy to mitigate bias caused by different task sizes. Results on VFN186-T2D and Food101-LT with $\rho=0.01$ are shown in Table~\ref{tab:large_small}.
Our method achieves the best performance on large tasks across both datasets, reaching $69.37\%$ on VFN186-T2D and $77.40\%$ on Food101-LT, indicating its strong ability to preserve knowledge from large tasks and prevent them from being disrupted by subsequent smaller ones. On small tasks, our method remains competitive with the best-performing approaches, suggesting that it can still effectively incorporate knowledge from small tasks without sacrificing the stability of larger ones. In addition, our method also performs well on middle-sized tasks, demonstrating a balanced ability to handle tasks of different scales under step-imbalanced continual learning.
\begin{table}[t]
\resizebox{0.9\columnwidth}{!}{%
\begin{tabular}{l|ccc|ccc}
\Xhline{0.8pt}
Datasets
& \multicolumn{3}{c|}{\textbf{VFN186-T2D}} 
& \multicolumn{3}{c}{\textbf{Food101-LT}} \\
Evaluation Metrics
& Large & Middle & Small
& Large & Middle & Small \\
\hline
SLCA~\cite{zhang2023slca} 
& 57.47 & 69.45 & 67.64 
& \underline{73.32} & 74.01 & 70.9 \\

InfLoRA~\cite{liang2024inflora}   
& 57.57 & 43.65 & 22.91
& 67.55 & 52.39 & 16.00 \\

EASE~\cite{zhou2024expandable}        
& 64.75 & 68.00 & 71.64 
& 59.96 &  77.89 & 81.4 \\

SEMA~\cite{wang2025self}              
& 42.87 & 44.76 & 31.27 
& 46.51 & 59.06 & 25.30 \\

CL-LoRA~\cite{he2025cl}               
& 49.85 & 57.93 & \textbf{76.36} 
& 46.45 & 70.67 & \textbf{72.7} \\

ACMap~\cite{fukuda2025adapter}        
& 63.18 & \underline{69.59} & 72.00 
& 46.45 & 70.67 & \textbf{72.7}\\

MOS~\cite{sun2025mos}   
& 65.57 & 68.76 & 65.45
& 72.36 & 79.05 & 63.90 \\

TUNA~\cite{wang2025integrating}      
& \underline{68.62} & 65.66 & 37.82
& 71.81 & \textbf{81.62} & 70.5 \\

\rowcolor{red!5}
DIME (Ours)
& \textbf{69.37} & \textbf{70.97} & \underline{73.09} 
& \textbf{77.40} & \underline{80.32} & 72.00 \\

\Xhline{0.8pt}
\end{tabular}
}
\centering
\vspace{-0.3cm}
\caption{Task-wise accuracy on large, middle, and small tasks on VFN186-T2D and Food101-LT. The best results are in \textbf{bold} and the second-best results are \underline{underlined}.}
\label{tab:large_small}
\vspace{-0.5cm}
\end{table}


\subsection{Ablation Study}
\label{sec:ablation}

\begin{table}[t]
\centering
\resizebox{\columnwidth}{!}{%
\begin{tabular}{lcccc|cc}
\hline
\textbf{Variant} & \textbf{SM} & \textbf{CCW} & \textbf{RTM} & \textbf{BSM} &
\multicolumn{2}{c}{\textbf{VFN186-Insulin}} \\
 & & & & & $A_T$ & $\mathrm{w}\bar{A}$ \\
\hline
Base             & (Direct Merge) & (Equal Weight) &  & (CE) & 66.73  & 74.90  \\
+ SM          & $\checkmark$ & (Equal Weight) &  & (CE) & 67.20 & 74.95  \\
+ CCW            & $\checkmark$ & $\checkmark$ &  & (CE) & 67.95   & 76.68  \\
+ RTM             & $\checkmark$ & $\checkmark$ & $\checkmark$ & (CE) & 68.68  & 77.67  \\
+ BSM            & $\checkmark$ & $\checkmark$ & $\checkmark$ & $\checkmark$ & \textbf{69.31}  & \textbf{78.07}  \\
\hline
\end{tabular}
}
\vspace{-0.3cm}
\caption{Step-by-step ablation study on VFN186-Insulin. The best results are in \textbf{bold}.}
\label{tab:ablation_step}
\vspace{-0.4cm}
\end{table}

We conduct ablation studies to analyze the contribution of each component in the proposed framework. 
Table~\ref{tab:ablation_step} reports step-by-step results on VFN186-Insulin.
Starting from the base configuration, incorporating spectral merging (SM) improves the performance from $66.73\%$ to $67.20\%$, indicating that aligning task adapters helps reduce conflicts across tasks. 
Adding class-count weighting (CCW) further boosts the performance to $67.95\%$, showing the benefit of accounting for step-level imbalance during adapter merging. 
Introducing rank-wise threshold modulation (RTM) brings an additional improvement to $68.68\%$, suggesting that selectively preserving dominant directions can stabilize knowledge integration. 
Finally, combining Balanced Softmax (BSM) addresses the long-tailed class distribution, achieving the best performance of $69.31\%$.
Overall, each component contributes to the performance improvement, demonstrating the effectiveness of the proposed framework for handling both step imbalance and class imbalance.

\subsection{Discussion}
\label{subsec:discussion}

\paragraph{Efficiency.}

\begin{table}[t]
\centering
\resizebox{0.82\columnwidth}{!}{%
\begin{tabular}{lccc}
\toprule
Method & Inference Time (s) $\downarrow$ & FLOPs (G) $\downarrow$ & Accuracy (\%) $\uparrow$ \\
\midrule
MOS        & 89.15 & 372.27 &  \underline{65.70} \\
ACMap      & 10.01 & \textbf{33.73}  & 64.52 \\
EASE       & 74.67 & 335.64 & 65.35 \\
SEMA       & \textbf{9.16} & 33.84  & 39.61 \\
CL-LoRA    & 80.90 & 335.82 & 49.87 \\
\rowcolor{red!5}
DIME (Ours) & \underline{9.50} & \textbf{33.73} & \textbf{68.49} \\
\bottomrule
\end{tabular}
}
\vspace{-3mm}
\caption{Efficiency comparison between adapter-based continual learning methods. 
The best results are in \textbf{bold} and the second-best results are \underline{underlined}.}
\vspace{-6mm}
\label{tab:efficiency}
\end{table}
Table~\ref{tab:efficiency} compares the inference efficiency of different adapter-based continual learning methods. DIME achieves the highest accuracy while maintaining inference cost comparable to lightweight approaches such as ACMap and SEMA. In contrast, several adapter-based methods including MOS, EASE, and CL-LoRA incur substantially higher inference cost due to the need to maintain multiple task-specific adapters. These results demonstrate that DIME achieves a favorable accuracy–efficiency trade-off, effectively improving performance without introducing additional inference overhead.

\vspace{-5mm}
\paragraph{Sensitivity.}

\begin{figure}[t]
    \centering
    \begin{subfigure}{0.56\linewidth}
        \includegraphics[width=\linewidth]{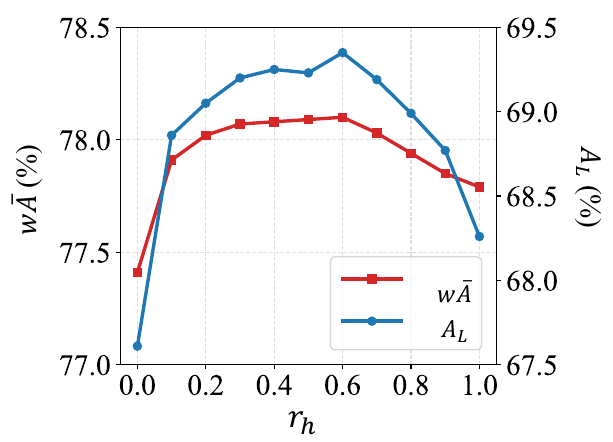}
        \caption{Sensitivity to the head-direction ratio  $r_h$.}
        \label{fig:gating_rh}
    \end{subfigure}
    \hfill
    \begin{subfigure}{0.42\linewidth}
        \includegraphics[width=\linewidth]{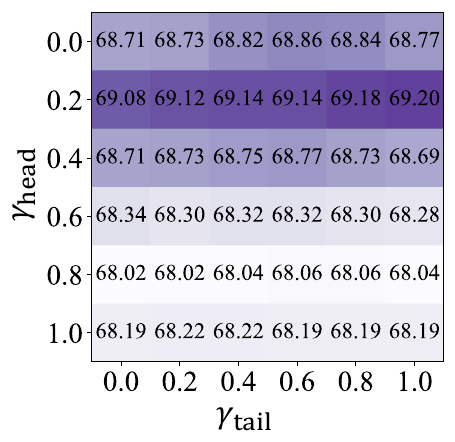}
        \caption{Sensitivity to the protection coefficients  $\gamma_{\text{head}}$ and $\gamma_{\text{tail}}$.}
        \label{fig:gating_gamma}
    \end{subfigure}
    \vspace{-0.2cm}
    \caption{Sensitivity analysis of rank-wise threshold parameters. 
    }
    \label{fig:gating_sensitivity_reduced}
    \vspace{-0.5cm}
\end{figure}

We analyze the sensitivity of the proposed threshold modulation mechanism to the head-direction ratio $r_{\mathrm{h}}$, which determines the proportion of singular directions treated as important. As shown in Figure~\ref{fig:gating_rh}, the performance first improves and then gradually saturates as $r_{\mathrm{h}}$ increases. When $r_{\mathrm{h}}$ is too small, only a few directions are protected, making the model more prone to forgetting previously learned knowledge. Increasing $r_{\mathrm{h}}$ helps preserve dominant directions and improves stability. However, overly large values lead to overly conservative updates, limiting the integration of new knowledge. These results suggest that a moderate head-direction ratio achieves the best balance between knowledge preservation and adaptation.

We further analyze the influence of the protection coefficients $\gamma_{\text{head}}$ and $\gamma_{\text{tail}}$. As shown in Figure~\ref{fig:gating_gamma}, the best performance is achieved when $\gamma_{\text{head}}$ is relatively small (around $0.2$), while larger values of $\gamma_{\text{tail}}$ generally lead to better results. This observation aligns with the intuition of rank-wise threshold modulation: head directions correspond to more important singular directions and should be updated more cautiously, whereas tail directions are less critical and can allow more flexible integration of new information. When $\gamma_{\text{head}}$ becomes too large, dominant directions become overly receptive to new information, which may disturb previously learned knowledge and slightly degrade performance. These results demonstrate that assigning smaller update strengths to head directions while allowing larger updates on tail directions helps balance knowledge preservation and adaptation.

Overall, these hyperparameters are not highly sensitive within a reasonable range. 
The specific values are provided in the released code for reproducibility.

\section{Conclusion}




In this work, we study continual food recognition under a more realistic dual-imbalanced learning scenario that closely reflects real-world dietary data streams, where food categories follow long-tailed distributions and the amount of data collected at different periods varies substantially. To address this challenge, we propose DIME, a parameter-efficient framework that integrates Balanced Softmax training with class-count aware spectral adapter merging and rank-wise threshold modulation. Extensive experiments on multiple long-tailed food benchmarks demonstrate consistent improvements over strong continual learning baselines while maintaining efficient inference with a single merged adapter. These results highlight the effectiveness of the proposed framework for robust and scalable food recognition in real-world dietary monitoring applications. Future work will explore extending the framework to more comprehensive food understanding tasks such as ingredient recognition and multi-label food recognition.
{
    \small
    \bibliographystyle{ieeenat_fullname}
    \bibliography{main}
}


\end{document}